\documentclass[journal]{IEEEtran}

\usepackage{color}
\usepackage[utf8]{inputenc} 
\usepackage[T1]{fontenc}    
\usepackage{hyperref}       
\usepackage{url}            
\usepackage{multirow}
\usepackage{tabularx}
\usepackage{booktabs}       
\usepackage{amsfonts}       
\usepackage{amsmath}       
\usepackage{nicefrac}       
\usepackage{microtype}      
\usepackage[pdftex]{graphicx}
\usepackage[switch]{lineno}

\newcommand\setrow[1]{\gdef\rowmac{#1}#1\ignorespaces}
\newcommand\clearrow{\global\let\rowmac\relax}
\clearrow

\newcommand{\mytilde}{\raise.2ex\hbox{$\scriptstyle\sim$}}

\definecolor{st}{rgb}{0.2118, 0.5059, 0.2235}
\newif\iffinal\finalfalse
\finaltrue

\iffinal
    \newcommand{\st}[1]{\ignorespaces}
    \newcommand{\panos}[1]{\ignorespaces}
    \newcommand{\todo}[1]{\ignorespaces}
\else
    \newcommand{\st}[1]{{\color{st}{\emph{#1}}}}
    \newcommand{\panos}[1]{{\color{red}{\emph{#1}}}}
    \newcommand{\todo}[1]{{\color{blue}{\emph{#1}}}}
\fi

\title{Unsupervised Online Multitask Learning of Behavioral Sentence Embeddings}

\author{
  Shao-Yen Tseng and 
  Brian Baucom and
  Panayiotis Georgiou, \IEEEmembership{Senior Member, IEEE}
}

\begin{document}
\maketitle

\begin{abstract} \panos{Focus is on robustness and domain portability through unsupervised training. Embeddings are just the tool so then need to be mentioned much less in abstract}
Unsupervised learning has been an attractive method for easily deriving meaningful data representations from vast amounts of unlabeled data.
These representations, or embeddings, often yield superior results in many tasks, whether used directly or as features in subsequent training stages. 
However, the quality of the embeddings is highly dependent on the assumed knowledge in the unlabeled data and how the system extracts information without supervision.
Domain portability is also very limited in unsupervised learning, often requiring re-training on other in-domain corpora to achieve robustness. 
In this work we present a multitask paradigm for unsupervised contextual learning of behavioral interactions which addresses unsupervised domain adaption. 
We introduce an online multitask objective into unsupervised learning and show that sentence embeddings generated through this process increases performance of affective tasks.


\end{abstract}

\section{Introduction}

\panos{same as abstract. Focus on unsupervised learning: w2v, seq2seq etc are all unsupervised learnings based on context. w2v exploits language use for word representations, seq2seq exploits interactions for interaction(chatbot), contextual (sentences) and behavioral (sequential chunks of multiple words) representations. These are all unsupervised contextual representation learning methods. We are focusing on  Unsupervised Behavioral Interaction Contextual Learning (or similar term)}

Representation learning has become a crucial tool for obtaining superior results in many machine learning tasks \cite{bengio2013representation}.
In the scope of natural language processing (NLP) a notable example of transforming input into more informative non-linear abstractions is word embeddings, or \textit{word2vec} \cite{Mikolov2013Efficient-Estim}.
Word embeddings exploit the use of language by learning semantic regularities based on a context of neighboring words.
This form of contextual learning is unsupervised, which allows learning from large-scale corpora and is a main reason for its strength and effectiveness in improving performance of many tasks such as constituency parsing \cite{tai2015improved}, 
 sentiment analysis \cite{dos2014deep, severyn2015twitter}, natural language inference \cite{parikh2016decomposable}, and video/image captioning \cite{karpathy2015deep, venugopalan2016improving}.

Later, with the introduction of sequence-to-sequence models \cite{sutskever2014_sequence-to-seq}, embeddings were extended to encode entire sentences and allowed representation of higher levels of concept through transformation of longer contexts. 
For example, \cite{kiros2015_skip-thought-ve} obtained sentence embeddings, which they referred to as \emph{skip-thoughts}, by training models to generate the surrounding sentences of extracts from contiguous pieces of text from novels. 
The authors showed that the embeddings were adept at representing the semantic and syntactic properties of sentences through evaluation on various semantic related tasks. 
In \cite{palangi2016_deep-sentence-e} the authors extracted sentence embeddings from an LSTM-RNN which was trained using user click-through data logged from a web search engine. 
They then showed that embeddings generated by their models were especially useful for web document retrieval tasks. 
Later, \cite{tseng2017_approaching-hum} extracted sentence embeddings from a conversation model and showed the richness of semantic content by applying an additional weakly-supervised architecture to estimate the behavioral ratings of couples therapy sessions.
Many other works have focused on obtaining general purpose sentence representations: sentence embeddings that are adept at multiple NLP tasks \cite{hill2016learning, conneau2017Supervised, cer2018universal}.


The benefit of many of the methods in aforementioned works is that the embedding transformation is learned on large amounts of unlabeled data.
Since natural language is an extremely complex process, it is crucial to leverage large corpora when learning embeddings so as to capture \textit{true} semantic concepts instead of regularities of the data, \textit{e.g.} domain-specific topics \cite{klein2005unsupervised}. 
Unsupervised learning allows us to utilize as much data as possible to increase the breadth of language understanding while minimizing the effort of data annotation.


However, a common issue with unsupervised training of sentence embeddings is the unpredictability of the resulting embedding transformation. 
In other words the embedding distribution is highly random and often contains redundant or irrelevant information. 
In addition, depending on training conditions such as architecture or dataset, it might fail to capture informational concepts or even semantics of the input data \cite{conneau2017Supervised}.
This is to be expected since the amount of information increases significantly as we move from words to sentences. 
It has been also noted that the quality of sentence embeddings is often highly dependent on the training dataset \cite{palangi2016_deep-sentence-e, tseng2017_approaching-hum}.
So much so that the use of embeddings trained on small domain-relevant datasets could yield results better than those trained on larger generic unsupervised datasets \cite{kiros2015_skip-thought-ve}. 

In this work we propose an online multitask learning (MTL) framework which aims to \textit{guide} unsupervised sentence embeddings into a space that is more discriminative in a final task. 
In our framework, transfer of domain-knowledge is achieved through an additional task in parallel with unsupervised contextual learning. 
The labels for the multitask are generated online from the unlabeled data to maintain the low annotation effort of an unsupervised scenario. 
Finally we apply the sentence embeddings to a final task of annotating human behaviors as evaluation and show improvement in the potency of unsupervised contextual learning through MTL. 



\section{Related work}

Many works have focused on leveraging multitask learning to enhance the informational content of sentence embeddings.
These methods can generally be categorized into task-specific or general-purpose applications.

In task-specific implementations a multitask function is often added to a primary supervised objective.
For example, \cite{yu2016_learning-senten} jointly learned sentence embeddings with an additional pivot prediction task in conjunction with sentiment classification. 
\cite{rei2017semi} predicted neighboring words as a secondary objective to improve accuracy of various sequence labeling tasks.

On the other hand, general purpose sentence embeddings aim to provide pre-trained features which, when transferred to unrelated tasks, improves overall performance.
\cite{luong2016_multi-task-sequ} achieved this by combining various tasks such as machine translation, constituency parsing, and image caption generation, which improved the translation quality between English and German.
Recently, \cite{subramanian2018learning} presented a large-scale multitask framework for learning general purpose sentence embeddings by training with a multitude of NLP tasks, including \textit{skip-thought} training, machine translation, entailment classification, and constituent parsing.
Similarly, universal sentence representations were also proposed in \cite{conneau2017Supervised} and \cite{cer2018universal}. 
\cite{conneau2017Supervised} used a single Natural Language Inference (NLI) task as the training objective whereas \cite{cer2018universal} also included tasks such as \textit{skip-thought} and response-generation. 

Our work differs in that we build on contextual learning and attempt to \textit{guide} unsupervised learning through a related multitask objective.
Unlike prior works, we target unsupervised scenarios and instead use a simple scheme to generate multitask labels online. 
Although unsupervised learning has historically required more data and training time, recent implementations of general purpose sentence embeddings have greatly scaled up training in both dataset size and model complexity. 
We show that through multitask \textit{guidance} unsupervised sentence embeddings can still excel in targeted tasks without requiring extensive labeled datasets or complicated models. 


In this paper we evaluate the performance of the unsupervised multitask sentence embeddings in identifying various human behaviors exhibited in conversational dialogue. 
In order to assess different sentence embedding methods fairly we apply simple machine learning techniques to obtain results for the final task rather than neural networks which would be able to exploit minor gains in the features.
We then provide an analysis of the results to give insight on the benefits of our proposed framework. 






\section{Unsupervised Multitask Embeddings }
\label{sec:embeddings}

%

\begin{figure*}[th]
	
	\centering
    \includegraphics[width=0.65\linewidth]{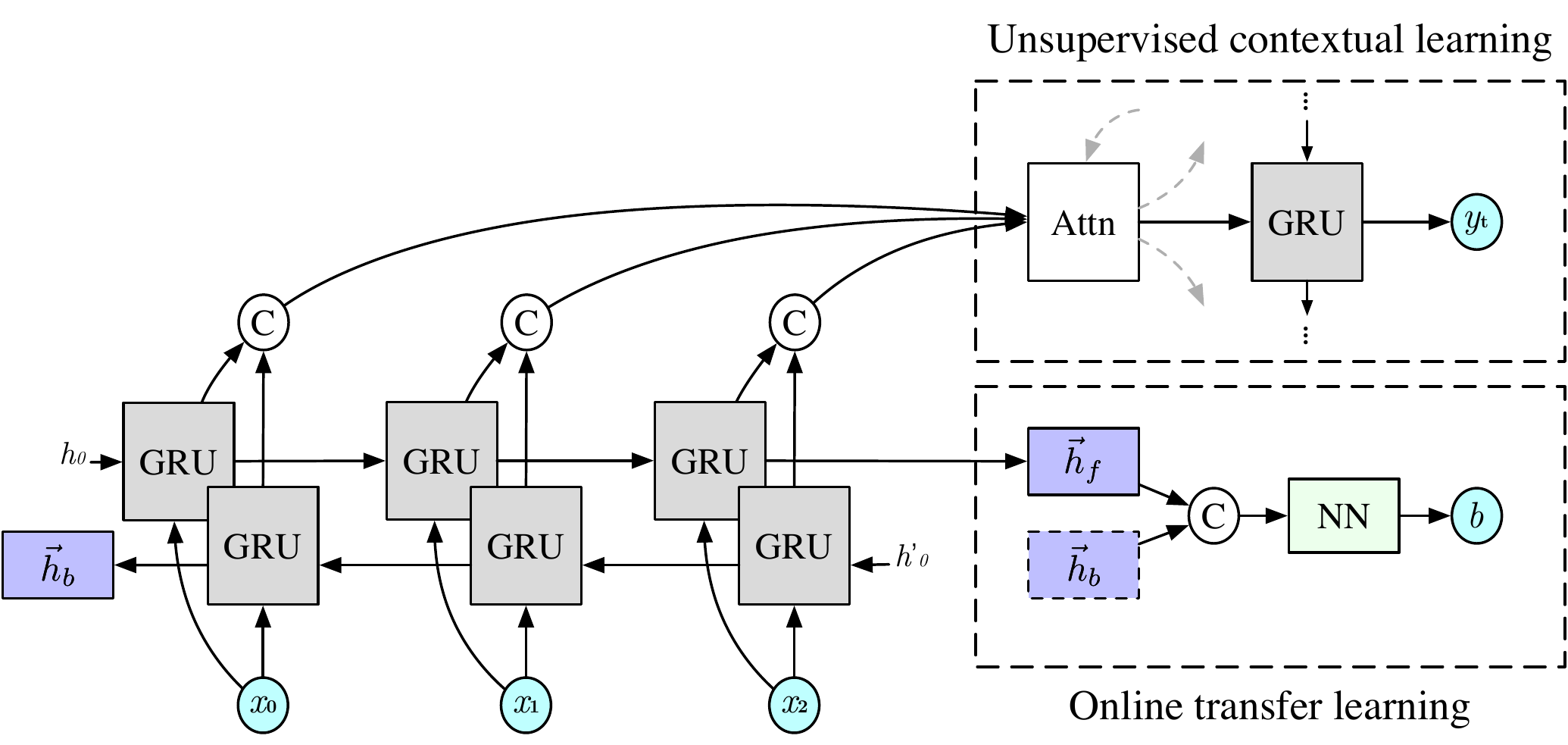}
    \caption{
     Bidirectional sequence-to-sequence conversation model with multitask objective. 
     The \texttt{GRU} blocks represent multi-layered RNNs using GRU units, \texttt{C} is the
     	concatenation function, and \texttt{Attn} is an attention mechanism.
     }
    \label{fig:mtseq2seq}
\end{figure*}

\subsection{Sequence-to-sequence sentence embeddings}

\st{Should we compare with Word2Vec and highlight benefits of sentence embeddings?}
\panos{yes. I think we do need to compare with w2v. On eval we also need to do binary \& interannotator agreement eval... build-up: w2v embeddings = local context, seq2seq=longer context; MTE=longer context \& Unsupervised 'behavior' context (need better term).}

The sequence-to-sequence model (seq2seq) \cite{sutskever2014_sequence-to-seq} maps input sequences to output sequences using an encoder-decoder architecture. 
Given an input sentence $ \mathbf{x} = (x_0, x_2, ..., x_T) $ and output sentence $ \mathbf{y}
= (y_0, y_2, ..., y_{T'}) $, where $ x_t $ and $ y_t $ represent individual words, the standard sequence model can be expressed as computing the conditional probability
\begin{equation}
\label{eq:y}
		P ( \mathbf{y} \mathbin{\vert} \mathbf{x} ) = 
			\prod_{t=0}^{T'} P( y_t \mathbin{\vert} y_{i<t}, \mathbf{s}, h)
\end{equation}
where $ \mathbf{s} $ is the sequence of outputs $ s_{t} $ from the encoder and $ h $ is the internal representation of the input given by the last hidden state of the encoder. 
For a given dataset $ \mathcal{D} = \{(\mathbf{x}_n,\mathbf{y}_n)\}_{n=1}^{N}$, the internal representation $ h $ can be expressed as
\begin{equation}
		h_{\theta} \equiv f(\mathbf{x} \mathbin{\vert} \mathcal{D}) = f(\mathbf{x} \mathbin{\vert}
		\boldsymbol\theta) 
\end{equation}
where $ f(\cdot) $ is the encoder function and $ \boldsymbol{\theta} $ is the set of parameters resulting from $ \mathcal{D} $. 

The internal representation $ h_{\boldsymbol{\theta}} $ encodes the input $ \mathbf{x} $ into an internal representation that allows the decoder to generate the best estimate of $\mathbf{y}$. 
In cases where $ \mathcal{D} $ contains semantically-related data pairs, $ h_{\boldsymbol\theta} $ can be viewed as a semantic vector representation of the input, or \emph{sentence embedding}, which can be useful for subsequent NLP tasks. 
In our case we apply contextual learning and designate consecutive sentences in continuous corpora as  $ \mathbf{x} $ and $ \mathbf{y} $.

While this model allows us to obtain semantic rich embeddings through training on unsupervised data, the quality of the embeddings is highly influenced by biases in the data and prevents the embeddings from becoming specialized in any target task \cite{conneau2017Supervised}. 
Therefore we propose to enhance the quality of unsupervised sentence embeddings through multitask learning.

\subsection{Multitask embedding training}

The addition of a multitask objective can guide embeddings into a space that is more discriminative in a target application.  
We hypothesize that this holds true even when the multitask labels are generated online from unsupervised data with no assumption of label reliability, as long as there is some relation between the multitask and target application. 

Assuming an online system which generates multitask labels $ \mathbf{b} $ for each input $ \mathbf{x} $ we can augment the dataset to yield 
$ \mathcal{D}_{\mathrm{aug}} = \{(\mathbf{x}_n,\mathbf{y}_n, \mathbf{b}_n)\}_{n=1}^{N}$.  
We then aim to predict this new label $ \mathbf{b} $ in conjunction with the original output sequence $ \mathbf{y} $. 
This is implemented in our seq2seq model by adding another head, or multitask network, after the internal representation $h$, as shown in Figure \ref{fig:mtseq2seq}.  
In addition to Eq. \ref{eq:y}, the model now also estimates the conditional probability
\begin{equation}
		P ( \mathbf{b} \mathbin{\vert} \mathbf{x} ) = 
			g( h \mathbin{\vert} \mathcal{D_{\mathrm{aug}}}) = g(h_{\theta_{\mathrm{aug}}})
\end{equation} 
where $ g(\cdot) $ is the multitask network and $ h_{\theta_{\mathrm{aug}}} $ is the new internal representation given by $ \mathcal{D}_{\mathrm{aug}} $.
In this work, the multitask network $ g(\cdot) $ is implemented with a multilayer perceptron. 

The training loss is then the weighted sum of losses from the multiple tasks, defined as
\begin{equation}
		J = 
		    \lambda \cdot L_1(\mathbf{y} , \mathbf{x} )+ 
			(1 - \lambda) \cdot L_2(\mathbf{b} , \mathbf{x})
\end{equation} 
where $ L_1 $ and $ L_2 $ are the cross entropy losses for contextual learning and the additional task, respectively.  

With most multitask setups there is an issue on how to control the training ratio $ \lambda $ to account for different data sources.
For example, if there is no overlap in inputs of the multiple tasks then $ \lambda $ can only alternate between 0 and 1 during training to switch between the different tasks. 
However, since we propose a multitask objective whose labels are generated from incoming data we are able to freely adjust $ \lambda $. 
It is possible to adjust the multitask ratio as training progresses to put emphasis on different tasks but we do not make any assumptions on the optimal weighting scheme and give equal importance to both tasks by setting $ \lambda $ to 0.5.


\subsection{Online multitask label generation}

To \textit{guide} the embeddings in becoming more human-behaviorally relevant, we select a multitask objective that attempts to classify the affective state of input sentences.
The definition of human behavior is more complicated than these states, however we hypothesize this is a suitable method of transferring related domain knowledge into the unsupervised sentence embeddings. 

We generate the affective labels for each input during training using an online mechanism. 
In our online approach we apply the simplest method by automatically labeling inputs using a simple look-up table of affective words \cite{tausczik2010_the-psychologic}.
Specifically, we use words categorized in the two top-level affective states: negative and positive emotion.  
An input sentence is assigned a \textit{Negative} or \textit{Positive} label based on the count of words corresponding to each affective state.
Although this labeling approach is extremely naive with a high rate of misclassification, we hypothesize the inclusion of affective knowledge in embeddings will be beneficial in identifying more complex behaviors or emotions later. 
Some examples of affective words in our look-up table are shown in Table \ref{tab:lookup}.

\begin{table}[htb]
  \caption{Examples of Positive and Negative Affect Words}
  \label{tab:lookup}
  \centering
  \begin{tabular}{>{\rowmac}l>{\rowmac}l>{\rowmac}l>{\rowmac}l>{\rowmac}l<{\clearrow}}
    \toprule
    \multicolumn{5}{c}{Affective State}                   \\
    \midrule
    Positive   & & \ \  &  Negative &  \\
    \midrule
    \setrow{\itshape} cute & love & & ugly & hate \\
	\setrow{\itshape} rich & nice & & hurt & nasty \\
	\setrow{\itshape} special &  sweet & & wicked & distraught \\
	\setrow{\itshape} forgive &  handsome & & shame & overwhelm \\
    \bottomrule
  \end{tabular}
\end{table}






\section{Behavior Identification using Embeddings}
\label{sec:behavior}
\st{For application should we use a more general tone or just aim for behavior annotation}
\panos{I will do a pass of language later and see the details. However this subsection should be a different section -- this isn't relating to context but to labeling}

After unsupervised multitask training the encoder in the seq2seq model is used to extract embeddings for use as features in behavior identification in long pieces of text (which we refer to as sessions). 
We define sentence embeddings to be the concatenation of the final output states of both the forward and backward RNNs in the encoder. 
We also concatenated the output states from all the intermediate layers of the encoder. 
This is an extension of history-of-word embeddings \cite{huang2018_fusionnet:-fusi} and is
motivated by the intuition that intermediate layers represent different levels of concept. 
By utilizing intermediate representations of the sentence, we hypothesize that more information related to human behavior can be captured. 
Annotation of human behavior using sentence embeddings was then applied using various unsupervised and supervised methods.

\subsection{Unsupervised clustering}

As an initial step we analyzed the performance of the embeddings on a behavior classification task without any supervision. 
We applied a simple k-means clustering method on individual sentence embeddings to obtain multiple clusters.
We then labeled the clusters by randomly selecting a single seed session and assigning the session label to the centroid which the majority of embeddings in the session were closest to.
During evaluation, session labels were predicted based on the centroid which the majority of embeddings from the session were closest to. 
\st{Description might be a bit complicated. Do we need a figure illustrating this?}

\subsection{k-Nearest neighbors}

For supervised classification we applied a simple k-nearest neighbor (k-NN) approach.
In k-NN, an embedding is labeled according to its k-nearest neighbors in the training set. 
The final session label was then obtained by a majority vote over all embeddings in the session.





\subsection{Rating estimation using neural networks}

Finally, we applied a neural network on top of the embeddings to estimate actual behavior ratings.
For this section we applied the framework proposed in \cite{tseng2017_approaching-hum}.
Sessions were segmented into sentences and represented as a sequence of embeddings. 
A sliding window of size 3 was applied over the embeddings followed by an RNN using LSTM units. 
The RNN was trained to predict the session rating from each window. 
The final session label was obtained by training a Support Vector Regressor to map from the median of the window predictions to the session rating. 
For more details the reader can refer to \cite{tseng2017_approaching-hum}.

%
%

\section{Experimental setup}
\label{sec:setup}

\subsection{Datasets}

\subsubsection{OpenSubtitles}
We used separate datasets to train the unsupervised and supervised portions of our proposed
method. 
Since our final task is behavior annotation of human interaction, we wish to use a dataset that contains conversational speech when learning the unsupervised sentence embedding. 
A natural choice for a source rich in dialogue is movie subtitles. 
To this end we used the OpenSubtitles Corpus \cite{tiedemann2009_news-from-opus-}.
This corpus was generated using data from the website \texttt{opensubtitles.org} and contains user-submitted subtitles of movies and TV shows. 

We applied additional pre-processing in addition to the steps already taken in \cite{tiedemann2009_news-from-opus-}. 
Mainly, we attempted to generate a back-and-forth conversation by taking consecutive lines in the subtitles and assigning them as utterance and replies in an interaction. 
As there is no speaker information in the corpus it is hard to distinguish between dialogues and monologues without the use of advanced content analysis methods. 
However, we assume that this difference in conversational continuity will be dampened by the large amount of data available. 
We also assume that monologues also represent some form of internal dialogue which closely ties with the concepts between sentences. 

Finally, we applied standard text processing techniques to clean up the text further. 
These included auto-correction of commonly misspelled words, contraction removal, and replacement of proper nouns through parts-of-speech tagging.
The final unsupervised training set consists of 30 million sentence pairs. 

\subsubsection{Couples Therapy Corpus}
We applied our unsupervised sentence embeddings to the task of annotating behaviors in human interactions. 
For this we used data from the UCLA/UW Couple Therapy Research Project \cite{Christensen2004Traditional-ver} which contains recordings of 134 real couples with marital issues interacting over multiple sessions.
In each session the couples each discussed a self-selected topic for around 10 minutes. 
The recordings of the session were then rated by multiple annotators based on the Couples Interaction \cite{Heavey2002Couples-interac} and Social Support \cite{Jones1998Couples-interac} Rating Systems. 
This rating system describes 33 behavioral codes rated on a Likert scale of 1 to 9, where 1 indicates strong absence and 9 indicates strong presence of the given behavior. 
The number of annotators per session ranged from 2 to 12, however the majority of sessions (\mytilde 90\%) had 3 to 4 annotators. 
Annotator ratings were then averaged to obtain a 33 dimensional vector of behavior ratings per interlocutor for every session.
The ratings were binarized to produce labels for the classification task and the Likert scale values were used for behavior rating estimation. 

In this work we focused on the behaviors \textit{Acceptance}, \textit{Blame}, \textit{Humor}, \textit{Sadness}, \textit{Negativity}, and \textit{Positivity}. 
Similar to prior works (\cite{chakravarthula2015_a-language-base, tseng2017_approaching-hum}) we used only the top and bottom 20\% of the dataset in terms of averaged behavior ratings. 
To train our models the dataset was split into train and test sets using a leave-one-couple-out scheme. 
That is, for each fold, one couple was used as the test set and the remaining as the train set.
This resulted in 85-fold cross-validation.

\subsubsection{IEMOCAP}
We also evaluated the effectiveness of our sentence embeddings in emotion recognition using the Interactive Emotional Dyadic Motion Capture Database (IEMOCAP) \cite{busso2008iemocap}.
This dataset contains recordings from five male-female pairs of actors performing both scripted and improvised dyadic interactions. 
Utterances from the interactions were then rated by multiple annotators for dimensional and categorical emotions. 
Similar to other works \cite{fayek2017evaluating, cho2018deep}, we focused on four categorical labels where there was majority agreement between annotators: \textit{happiness}, \textit{sadness}, \textit{anger}, and \textit{neutral}, with \textit{excitement} considered as \textit{happiness}.
We used the transcripts from the dataset and removed any acoustic annotations such as laughter or breathing.  
After discarding empty sentences our final dataset consisted of 5,500 utterances (1103 for \textit{anger}, 1078 for \textit{sadness}, 1615 for \textit{happiness}, and 1704 for \textit{neutral}).  
To train the supervised layers we used leave-one-pair-out  which resulted in a 5-fold cross-validation scheme.

\subsection{Model architectures and training details}

\subsubsection{Sentence embeddings}

The sequence-to-sequence model with multitask objective, shown in Figure \ref{fig:mtseq2seq},
can be described as three sections: the encoder, the decoder, and the multitask network. 
The encoder was constructed using a multi-layered bidirectional RNN using GRU units. 
We performed a grid search using hyper-parameter settings of 2 and 3 layers, and, 100 and 300 dimensions in each direction per layer. 
For the decoder a unidirectional RNN using GRU units was used instead of bidirectional. 
The number of layers in the decoder were the same as the encoder while the dimension size was doubled to account for the concatenation of states and outputs from both directions. 

The multitask network was implemented using a neural network with four hidden layers of sizes 512, 512, 256, and 128.
We used rectified linear unit (ReLU) function as activation functions in the hidden layers and 2-dimensional softmax before the final output. 
No other network hyper-parameters were tried for the multitask network.

The sentence embedding models were trained with the OpenSubtitles dataset for 5 epochs using SGD with momentum. 
The learning rate was set to 0.05 and momentum set to 0.9.
We also reduced the learning rate by a factor of 10 every epoch.

\subsubsection{Supervised behavior annotation}
Similar to \cite{tseng2017_approaching-hum} we used an RNN with LSTM units to estimate behavior ratings in the Couples Thearpy Corpus. 
The RNN had a single layer with a dimension size of 50 in the LSTM unit. 
A sigmoid function was applied before the output to estimate the normalized rating value. 
In each fold one couple was randomly selected as validation to select the best model. 

\panos{how about multiple experts?}

\subsubsection{Supervised emotion recognition}

A neural network with four hidden layers was used to classify emotions using embeddings of sentences from the IEMOCAP dataset.  
The hidden layers were of size 256 and used ReLU as the activation function. 
The model was trained for 20 epochs using Adagrad \cite{duchi2011adaptive} as the optimization method. 
No other network hyper-parameters were tried for the emotion recognition network.
A subset of the training data (\mytilde 10\%) was used as validation in selecting the best model. 

\begin{table*}[ht]
  \centering
  \caption{\label{tab:results} Behavior Identification Accuracy (\%) using Multitask Sentence Embeddings}
  \begin{tabular}{llccccccc}
    \toprule
    Method & Embedding Model & Acceptance & Blame & Negativity & Positivity & Sadness & Humor & Mean Accuracy  \\
    \midrule
    k-Means & InferSent \cite{conneau2017Supervised} & 58.9 & 63.6 & 61.4 & 62.1 & 58.9 & 60.7 & 60.93\\
    & GenSen \cite{subramanian2018learning} & 53.9 & 66.4 & 61.4 & 61.4 & 59.6 & 58.9 & 60.27 \\
    & Universal Sentence Encoder \cite{cer2018universal} & 59.3 & 65.7 & 61.8 & 64.3 & 59.6 & 59.6 & 61.72 \\
    & Conversation Model \cite{tseng2017_approaching-hum} & 61.9 & 65.4 & 64.6 & \textbf{65.7} & 57.9 & 59.1 & 62.43 \\ 
    & \ \ + Online MTL (proposed) & \textbf{64.0}   & \textbf{66.4} & \textbf{65.0} & 62.1 & \textbf{61.4} & \textbf{62.1} & \textbf{63.50}\\
     
    \midrule
    k-NN & InferSent \cite{conneau2017Supervised} & 83.2 & 81.1 & 85.4 & 78.6 & 65.7 & 57.1 & 75.27 \\
    & GenSen \cite{subramanian2018learning} & \textbf{85.0} & 85.0 & 85.7 & 81.1 & 63.2 & 56.1 & 76.02    \\
    & Universal Sentence Encoder \cite{cer2018universal} & 80.0 & 82.5 & 83.9 & 79.6 & 66.8 & \textbf{60.4} & 75.53 \\
    & Conversation Model \cite{tseng2017_approaching-hum} & 79.6 & 80.0 & 85.7 & 82.5 & 64.6 & 59.6 & 75.53 \\
    & \ \ + Online MTL (proposed) & \textbf{85.0} & \textbf{85.4} & \textbf{87.9} & \textbf{86.8} & \textbf{67.9} & 60.0 & \textbf{78.77} \\
    \bottomrule
  \end{tabular}
\end{table*}

\begin{table}[ht]
    \centering
    \caption{Weighted Accuracy of Emotion Recognition on IEMOCAP}
    \label{tab:iemocap}
    \begin{tabular}{lc}
        \toprule
        Method & WA (\%) \\ 
        \midrule
        
        Lex-eVector \cite{jin2015speech} & 57.40  \\
        E-vector + MCNN \cite{cho2018deep} & 59.63  \\
        
        mLRF \cite{gamage2017salience} & 63.80 \\
        
        \midrule
        InferSent \cite{conneau2017Supervised} + DNN & 62.60 \\ 
        GenSen \cite{subramanian2018learning} + DNN & 60.62 \\ 
        Universal Sentence Encoder \cite{cer2018universal} + DNN & \textbf{64.83} \\ 
        Conversation Model \cite{tseng2017_approaching-hum} + DNN & 55.82 \\ 
        \ \ + Online MTL (proposed) + DNN & 63.84 \\ 
        \bottomrule
    \end{tabular}
\end{table}



\section{Experimental Results}
\label{sec:results}

We compared our unsupervised multitask sentence embeddings to general purpose embeddings such as InferSent \cite{conneau2017Supervised}, GenSen \cite{subramanian2018learning}, and Universal Sentence Encoder \cite{cer2018universal}.
Table \ref{tab:results} shows the results of behavior identification using sentence embeddings for different behaviors in the Couple Therapy Corpus. 
The addition of the multitask objective improved the classification accuracy of unsupervised sentence embeddings from the conversation model across all behaviors except \textit{Positivity} in unsupervised classification with k-Means. 
Under supervised learning using k-NN, our multitask embeddings improved accuracy on all behaviors except \textit{Humor}. 
In terms of mean accuracy, our multitask embeddings performed better than other sentence embeddings with an absolute improvement over no multitasking of 1.07\% and 3.24\% for unsupervised and supervised methods respectively. 
Our multitask embeddings also achieved the highest mean accuracy over all the sentence embeddings tested.

\todo{Results on inter-annotator agreement.}


\panos{Need a table that lists, for each behavior and averages, the w2v, s2s, s2s-MT, and hopefully s2s-MT-ML (multitask and multilabel) together... e.g. (see in fixed font)\\
            w2v   s2s   s2s-MT  s2s-MT-ML\\
acceptance  \#     \# ...\\ 
blame\\
.\\
.\\
.\\
average\\
}

\panos{and again, these are for the binary task and for the rating using kappa as in your stockholm paper}

The results of emotion recognition on IEMOCAP are shown in Table \ref{tab:iemocap}.
In addition to general purpose embeddings we also compared with other works that only used transcripts (\cite{cho2018deep, jin2015speech,  gamage2017salience} ). 
It should be noted that there is no consensus on data split and evaluation conditions in IEMOCAP, and while we made every effort to be consistent with other works the results may not be directly comparable. 
However, when comparing among our implementation using sentence embeddings we observed that online MTL improved the weighted accuracy (WA) of unsupervised embeddings by an absolute value of 8.02\% which is more than 14\% relative improvement.  
The highest accuracy was obtained using embeddings from the Universal Sentence Encoder, however our implementation was a close second by less than one percent. 


\label{sec:discussion}
\panos{I expect it will be along the lines of unsupervised MT helps since it is introducing human knowledge. Multilabel and multitask helps more since there are common pieces of info among various labels, e.g. blame is often expressed through negativity}

\begin{figure}[ht]
  \centering
  \includegraphics[width=\linewidth]{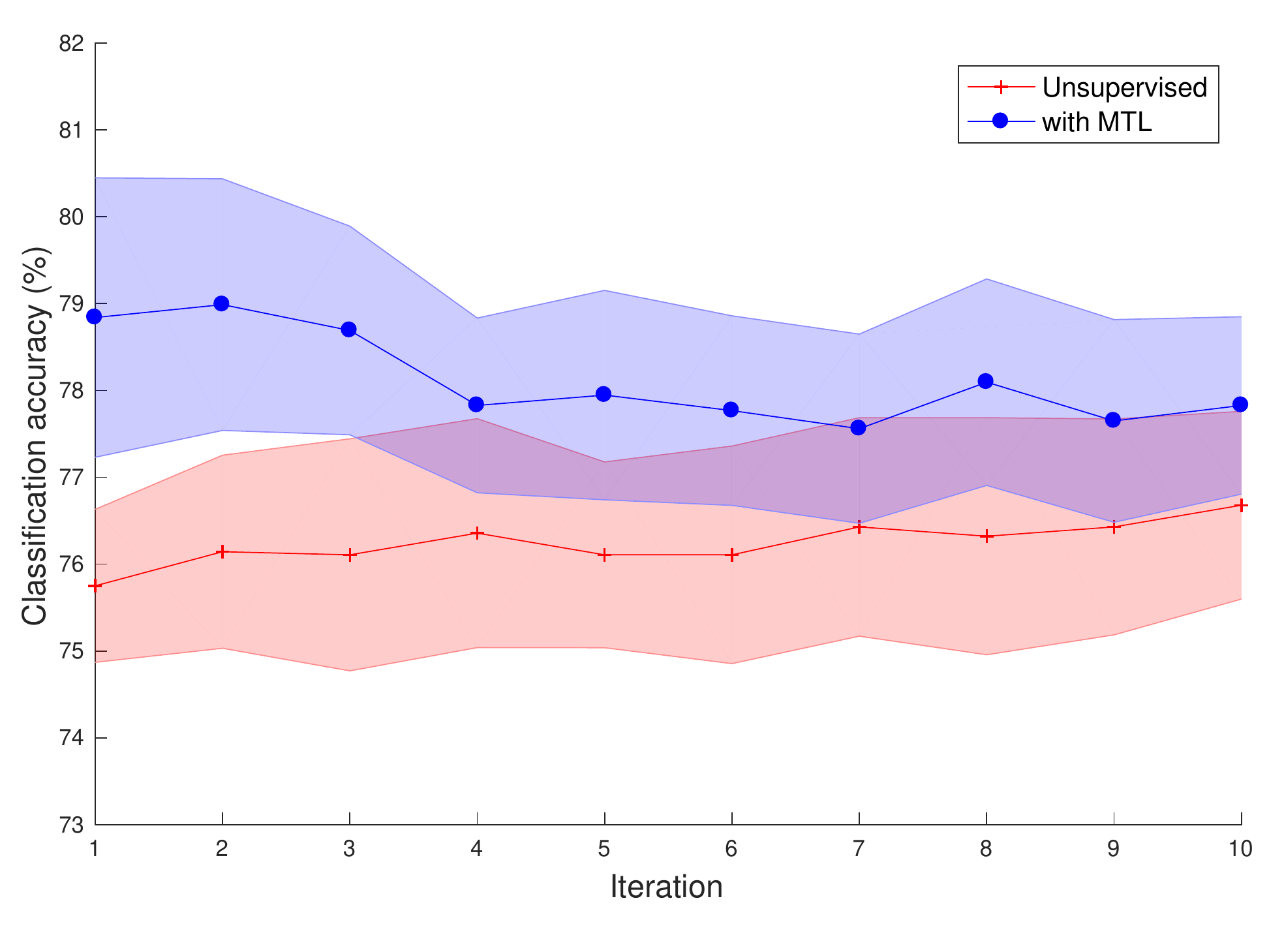}
  \caption{\label{fig:compare_acc} Standard error plot of classification accuracy on \textit{Negativity} across checkpoints for various model configurations. }
\end{figure}

Finally, we analyzed the performance of our sentence embeddings on \textit{Negativity} classification in behavior identification over the progression of training across different model architectures.
From the standard error plot, shown in Figure \ref{fig:compare_acc}, we can observe that the addition of the multitask learning objective collectively increases performance in the final task.
This shows that online transfer learning through multitask was successful at improving the performance of unsupervised sentence embeddings in our final task.

\panos{You need to simplify this story. Either chose a checkpoint, or chose a simple ensemble method (e.g. average the 3 closest in agreement), or use median... but this plot doesn't help... it's damaging the story}


\section{Conclusion}
\label{sec:conclusion}

In this work we explored the benefits of introducing additional objectives to unsupervised contextual learning of sentence embeddings. 
We found empirical evidence that supports the hypothesis that multitask learning can increase affective concepts in unsupervised sentence embeddings, even when the multitask labels are generated online and extremely unreliable. 
Our proposed model has the benefit of not requiring additional effort in generating or collecting data for multitasks. 
This allows learning from large-scale corpora in an unsupervised manner while simultaneously applying transfer learning. 
In contrast to general purpose sentence embeddings, our model learns sentence representations using less complex models and training effort, while at the same time yields higher performance in our target task. 
We argue that when learning sentence embeddings, it is more beneficial to apply \textit{guided} unsupervised learning instead of overemphasis on universality before domain transfer. 

While we do expect that further improvements can be obtained through better labels for the multitask objective, that would entail additional effort in system design and label generation. 
In addition, we also expect that multitask labels that are too domain-specific (e.g. focusing on a specific way or definition of affective expression) may actually hinder the performance of unsupervised embeddings. 
However, we do not verify this claim and leave it to future work.




\bibliographystyle{IEEEtran}
\bibliography{ref}

\end{document}